\documentclass[sigconf]{acmart}
\usepackage{multirow}
\usepackage{balance}
\usepackage{float}
\usepackage{caption}
\usepackage{subcaption}
\usepackage{CJKutf8}

\AtBeginDocument{%
  }

\copyrightyear{2023}
\acmYear{2023}
\setcopyright{acmlicensed}\acmConference[MM '23]{Proceedings of the 31st
ACM International Conference on Multimedia}{October 29-November 3,
2023}{Ottawa, ON, Canada}
\acmBooktitle{Proceedings of the 31st ACM International Conference on
Multimedia (MM '23), October 29-November 3, 2023, Ottawa, ON, Canada}
\acmPrice{15.00}
\acmDOI{10.1145/3581783.3611929}
\acmISBN{979-8-4007-0108-5/23/10}

\begin{document}
\begin{sloppypar}
\title{TextPainter: Multimodal Text Image
Generation with Visual-harmony and Text-comprehension for Poster Design}

\author{Yifan Gao}
\authornote{This work was done when Yifan Gao was at Alibaba as an intern.}
\orcid{0009-0001-4239-3418}
\affiliation{%
  \institution{University of Science and Technology of China}
  \city{Hefei}
  \country{China}
}
\email{eafn@mail.ustc.edu.cn}

\author{Jinpeng Lin}
\affiliation{%
  \institution{Alibaba Group}
  \city{Beijing}
  \country{China}
}
\email{linjinpeng.ljp@alibaba-inc.com}

\author{Min Zhou}
\affiliation{%
  \institution{Alibaba Group}
  \city{Beijing}
  \country{China}
}
\email{yunqi.zm@alibaba-inc.com}

\author{Chuanbin Liu}
\authornote{Corresponding author.}
\affiliation{%
  \institution{University of Science and Technology of China}
  \city{Hefei}
  \country{China}
}
\email{liucb92@ustc.edu.cn}

\author{Hongtao Xie}
\affiliation{%
  \institution{University of Science and Technology of China}
  \city{Hefei}
  \country{China}
}
\email{htxie@ustc.edu.cn}

\author{Tiezheng Ge}
\author{Yuning Jiang}
\affiliation{%
  \institution{Alibaba Group}
  \city{Beijing}
  \country{China}
}
\email{tiezheng.gtz@alibaba-inc.com}
\email{mengzhu.jyn@alibaba-inc.com}

\renewcommand{\shortauthors}{Yifan Gao, et al.}

\begin{abstract}
Text design is one of the most critical procedures in poster design, as it relies heavily on the creativity and expertise of humans to design text images considering the visual harmony and text-semantic.
This study introduces TextPainter, a novel multimodal approach that leverages contextual visual information and corresponding text semantics to generate text images. Specifically, TextPainter takes the global-local background image as a hint of style and guides the text image generation with visual harmony.
Furthermore, we leverage the language model and introduce a text comprehension module to achieve both sentence-level and word-level style variations.
Besides, we construct the PosterT80K dataset, consisting of about 80K posters annotated with sentence-level bounding boxes and text contents. We hope this dataset will pave the way for further research on multimodal text image generation.
Extensive quantitative and qualitative experiments demonstrate that TextPainter can generate visually-and-semantically-harmonious text images for posters.
\end{abstract}




\maketitle

\begin{figure}
\centering
\includegraphics[width=0.5\textwidth]{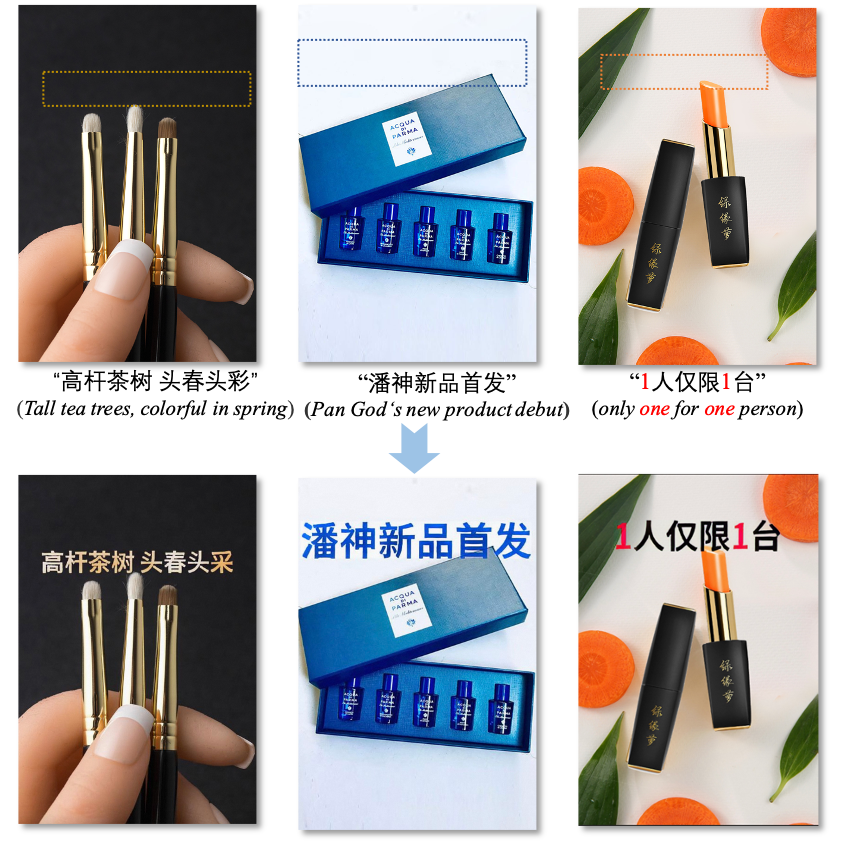}
\caption{Text image generation for poster design using TextPainter (English translation in brackets). (Top) Given a clean image without filled text, the content and position of the text. (Bottom) TextPainter generated harmonized text images. TextPainter is sensitive to the content of poster background image (left \& middle). TextPainter is capable of highlighting keywords (right).}
  \label{fig:main_img}
  \setlength{\abovecaptionskip}{0.cm}
  \vspace{-0.4cm}
\end{figure}

\section{Introduction}
\label{sec:intro}

Text design is an important subtask in poster design, where people create and render harmonious text for posters to convey information clearly and effectively. With the development of deep learning, researchers have made a series of successful attempt around poster design~\cite{ShunanGuo2021VinciAI,PraneethaVaddamanu2022HarmonizedBC}, e.g. layout generation~\cite{DBLP:conf/ijcai/ZhouXMGJX22,DBLP:conf/mm/CaoMZLXGJ22}, text content generation~\cite{DBLP:journals/corr/abs-2204-12974}. However, there has been little research conducted on text design.

In this paper, we propose a novel task entitled Text Image Generation for Posters (TIGER), which aims to accomplish text design with the automated method. Specifically, the goal is to generate text images for a specific line based on its position and text content. As demonstrated in Figure~\ref{fig:main_img}, generating high-quality text images which are clear, harmonious in color, and semantically appealing is a challenging task. 
Though this task seems to be similar to font generation~\cite{JiangYue2019SCFontSC, PengyuanLyu2017AutoEncoderGG, YimingGao2020GANBasedUC} or scene text generation tasks~\cite{LiangWu2019EditingTI, QiangpengYang2020SwapTextIB, krishnan2021textstylebrush}, it has three main unique and challenging features. 

Firstly, the generation style should not be explicitly predefined but implicitly relevant to the background. The goal of TIGER  is to create a visually pleasing text image by skillfully integrating the text content with the poster background in a natural, beautiful, and attractive manner. We aim to produce a distinctive visual effect that is not achieved by imitation of a predefined style. However, the font or scene text generation is to extract the predefined style as a reference and transfers it to other text images. For instance, the text colors in the first and second columns of Figure~\ref{fig:main_img} are consistent with their corresponding subjects and clearly distinguishable from the surrounding background regions.

Secondly, during the rendering of the generated text images, each character should be differently designed according to its semantics.
Comprehending the poster copy can effectively emphasize the keywords that are meant to be exhibited to the audience. 
As shown in the last column of Figure~\ref{fig:main_img}, the highlighted keywords, are differentiated from others through distinct styles.
In contrast, the font or scene text generation task usually demonstrates independent visual style and text semantics. 

Finally, the present absence of a properly structured and annotated dataset for this task exacerbates the challenge of training models. In particular, textual style attributes such as font, color, opacity, and outline present difficulties in manual annotation. Besides, image size and proportions vary and are not consistent.

To mitigate the aforementioned challenges, we propose a novel multimodal approach named TextPainter that exploits the contextual visual style of posters and the corresponding text semantics to produce text images that are both visually and linguistically meaningful.
Specifically, our approach initially approximates the overall color style of the text image by analyzing the global-local information of the poster background. Furthermore, it refines the fine-grained color of specific characters of the text image to match the text semantics at both the word and sentence levels, resulting in more innovative designs.

Firstly, we introduce a StyleGAN-based method for generating visually harmonious text images. Specificall, the text initially is rendered as an image, then encoded by the glyph encoder, which is used as input to the generator along with the background patch of the text area. Simultaneously, a color style encoder is utilized to extract the implicit global and local styles of the poster background and guide the generation process.

Secondly, a text understanding module is introduced to enhance the visual design ability of TextPainter through the utilization of text semantics. Specifically, we use Language-Image Pre-training (CLIP) models to extract text semantic tokens and then fuse word-level  and text-level tokens with visual features in various methods.

Finally, we employed image-level loss functions and several training techniques to facilitate the effective training of TextPainter. These approaches effectively address the issues of the dataset with weakly supervised annotations and variations in image size.

Besides, we constructed a dataset, called PosterT80K, to validate our proposed approach. It consists of 87,529 posters and 342,579 text elements collected from real-world use cases. Each element has been annotated with its bounding box and content. In particular, it was demonstrated through extensive experiments conducted on the dataset, which showed the effectiveness of our approach. Our main contributions can be summarized as follows:
\begin{itemize}
    \item We have proposed a new task, text image generation for poster design, which is aimed to generate clear, color harmony and creative text images pixel by pixel on poster backgrounds.
    \item We propose the TextPainter that utilizes the contextual visual style of the poster and the corresponding text semantics for text image generation, which is the first method to utilize text semantics to help text image generation to the best of our knowledge.
    \item We construct a large-scale poster dataset, Poster-T80K, consisting of 87,529 posters designed by designers with sentence-level bounding boxes and content of text string annotations.
\end{itemize}

\begin{figure*}[h]
   \setlength{\abovecaptionskip}{0.cm}
  \includegraphics[width=\textwidth]{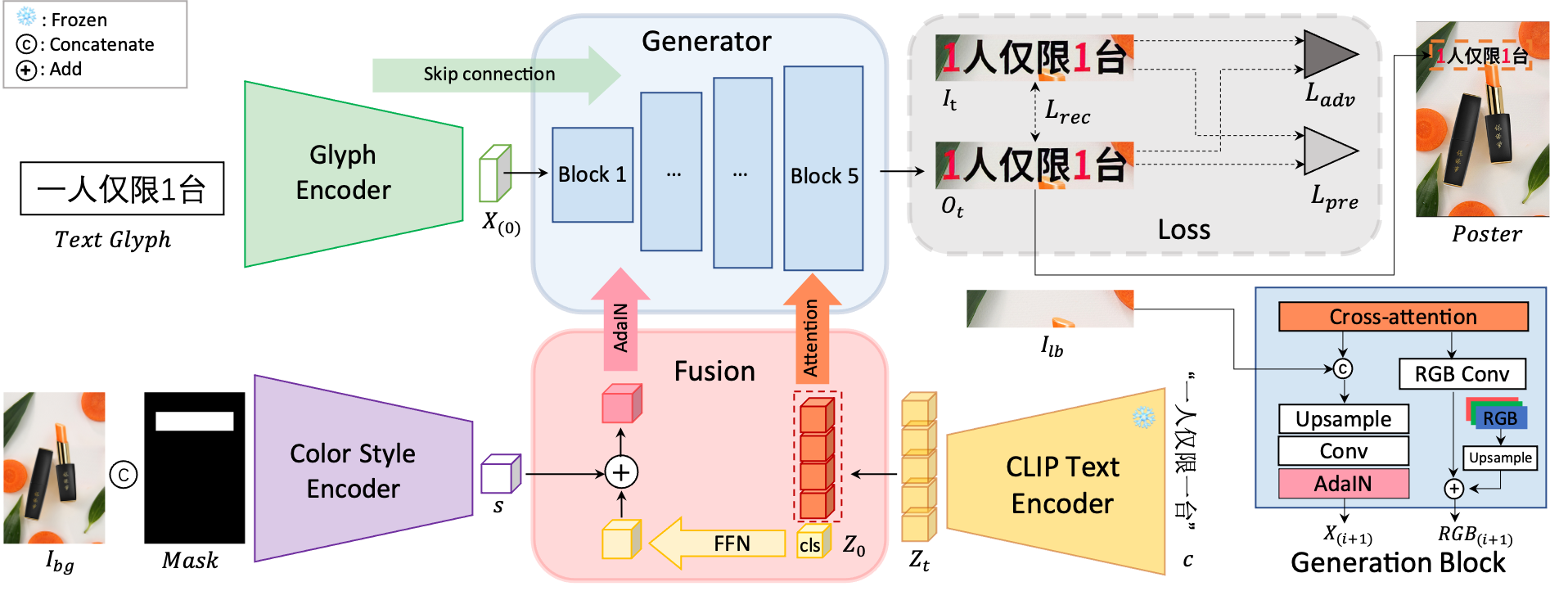}
  \caption{TextPainter contains five modules, glyph encoder to encode the text glyph, color style encoder to extract the image style features, CLIP text encoder to encode text to semantic tokens, fusion module to make the association between the text semantics and vision features, a generator to generate the text image.}
  \label{fig:net}
  \vspace{-0.4cm}
\end{figure*}

\section{Related Work}
\label{sec:related-work}
\subsection{Image-to-image Generation and Style Transfer}
With the advance of generative adversarial networks~\cite{IanGoodfellow2014GenerativeAN}, the quality of image generation is getting improved and the controllability of the generated content is getting easier. Pix2pix~\cite{PhillipIsola2016ImagetoImageTW} proposed a general solution to image-to-image generation problems using conditional adversarial networks. Wang et al.~\cite{TingChunWang2017HighResolutionIS} extended these efforts to high-resolution image generation. Style transfer can be seen as a special kind of image-to-image translation task, which modifies attributes of images, such as their style while keeping their content. Gatys et al.~\cite{gatys2015neural} published the first neural algorithm that creates artistic images of high perceptual quality by a pre-trained convolutional neural network. Using adaptive instance normalization~(AdaIN) that affines transformation parameters in normalization layers to represent styles. Huang et al.~\cite{XunHuang2017ArbitraryST} propose a method that is capable of real-time image style transfer. Zhu et al~\cite{zhu2017unpaired} proposed a cycle-consistent generative adversarial network to learn the one-to-one mapping of two domain images free from the dependence on paired training data.  MUNIT~\cite{XunHuang2018MultimodalUI} and DRIT~\cite{HsinYingLee2018DiverseIT} disentangled the representation of images into a domain-invariant content code and a domain-specific style code and can generate diverse outputs from a given source domain image. 

The series of work on style-based generators~\cite{TeroKarras2018ASG,LehtinenJaakko2019AnalyzingAI,TeroKarras2021AliasFreeGA} continues to break through the quality and stylistic diversity of generated images. Furthermore, this paper presents a novel text-image generation network that also leverages a style-based generator to achieve contextual harmony between images and text semantics.

\vspace{-0.4cm}
\subsection{Font Generation and Text Image Generation}
Font generation aims at transferring the typographic style of one font to another. Based on Pix2Pix, some early studies~\cite{JiangYue2019SCFontSC, PengyuanLyu2017AutoEncoderGG, YimingGao2020GANBasedUC} used paired data to train font generation network. Subsequent works~\cite{YangchenXie2021DGFontDG, YimingGao2018CalliGANUM} were able to perform unsupervised font generation. Another research direction is few-shot font generation (FFG)~\cite{YaoxiongHuang2020RDGANFC, SamanehAzadi2017MultiContentGF, YangchenXie2021DGFontDG,LichengTang2022FewShotFG,WeiLiu2022XMPFontSC}, where the transfer of an entire font can be accomplished with a few samples of the target font. 

Unlike the font generation task, style-guided text image generation takes into account not only typographic stylization but also textual stylization~(e.g., color and effect), which is more challenging. SRNet~\cite{LiangWu2019EditingTI} is the first attempt to edit the text in natural images on the word level with an end-to-end trainable style retention network. By predicting geometrical attributes of style images and the TPS~(Thin-Plate-Spline) module, SWAPText\cite{QiangpengYang2020SwapTextIB} is able to handle severe font geometric distortions. These methods require target-style images as supervision for model training and are therefore constrained by synthetic images. To tackle this problem, TextStyleBrush~\cite{krishnan2021textstylebrush} proposes a self-supervised one-shot text style transfer approach that can disentangle the style of a text image of all aspects of its appearance and shows impressive results of scene text content replacement. In contrast to TextStyleBrush, which is English character-based, TextPainter is used to handle text image generation of Chinese characters which has a more complex structure in general and is therefore more challenging. Besides, APRNet~\cite{shi2022aprnet} introduces a content-style cross-attention module and pixel sampling approach to achieve photo-realistic text image generation. 
Equipped with the Cross-Attention mechanism, TextPainter introduces the text comprehension module to build a bridge between text images and text semantics.

\subsection{Context-aware Text Image Generation}
Style-guided text image generation requires a target style image as a reference and imitates its appearance while keeping text content unaltered. Without using a target style image, some works try to model text image generation based on the context in which the text image is located. To aid the selection of fonts, colors, and sizes for designers in the process of designing web pages, Miyazono et al.~\cite{zhao2018modeling} propose to model the font in the context of web pages using multi-task deep neural network. This approach relies on structured HTML tags and models color as a discrete classification problem, resulting in insufficient color change capability. Similarly, in order to assist the text design for posters, we propose the generation of font images based on the poster context, but with freer pixel-level output space.

\section{Method}
\label{sec:method}

This section provides a brief overview of the TextPainter model's architecture, which is based on the StyleGAN framework. We present a methodology for text image generation that incorporates both local and global color harmony specifically designed for the poster design task. Additionally, we discuss how the text comprehension capability of existing multimodal models can enhance the visual design of text images. Lastly, we propose a specific padding method for image preprocessing to handle different image sizes within a single training batch.


\subsection{Overview}
We illustrate in Fig~\ref{fig:net} that TextPainter can automatically generate the text image $O_{t}$ on the poster by utilizing the poster's background image $I_{bg}$, text content $c$, and position $b$. When designing posters, color coordination between the local area of the text and the overall poster must be carefully considered, especially for text visual design. The process of generating a text image involves the color style encoder extracting information about both the local background of the text area $I_{lb}$ and the overall poster background $I_{bg}$, which is then integrated into the generator.

Moreover, we employ CLIP~\cite{chinese-clip}, a pre-training vision-language model, to extract text semantic information and enhance TextPainter's understanding of text semantics. This helps us create a visual design that highlights the key points of the text.

During the training phase, we adopt image-level supervision with the assumption that the presence of a single text in each image. A dataset was constructed by amassing a substantial collection of e-commerce-style poster images, and the text content and bounding boxes were labeled. In real-world applications, the text on the poster has different font sizes and lengths. In order to enable the model to perceive this information, we employ a padding method that uses background values during training.

\subsection{Text-Image Generation with Local and Global Color Harmony}
We use StyleGAN as the basis for our text image generation architecture, with some enhancements. To generate a text image on a poster background, we first need to render the text content into an image. Then, we use the glyph encoder to obtain the glyph feature map, which is used to generate the re-colored text image with the generator. However, this approach presents three challenges: 1) the downsampling of the initial text content image may cause the loss of structural information, 2) restoring the background of the text area based solely on the text content is an ill-posed problem, and 3) a method must be devised to ensure that the color scheme of the generated text image is harmonious with that of the poster.

To retain the glyph's structure, skip connections based on the U-net architecture are incorporated between the glyph encoder and the generator. Moreover, to enable the generator to focus only on generating the foreground text, the textual region's background image is provided as input. To address the final challenge of ensuring that the generated text images are harmonious with the poster background, several auxiliary loss functions are adopted to satisfy the constraints.

\subsubsection{Global image style awareness}
TextPainter employs a style encoder to encode the background image $I_{bg}$ into a background style vector $s$ to perceive its style. Owing to the hierarchical structure of the generator, distinct layers control styles at various levels. The style vector $s$ is mapped to multiple hierarchical style vectors $w_i$, and the AdaIN modules control $w_i$ to operate the style generator.

\subsubsection{Local patch awareness}
We previously mentioned the color style vector $s$, which fails to consider the position of the text. As a result, there is a possibility that text images generated at different positions may have identical colors, which is not desirable for poster visual design. To overcome this issue, we propose a solution that concatenates the poster background $I_{bg}$ and the position mask image along the channel dimension. This concatenated image is then inputted into the style encoder. By doing so, the style encoder can obtain style information that is conditioned on the text position, thereby ensuring that the generated text images have harmonious colors with the poster background.

\begin{figure}[h]
  \includegraphics[width=0.5\textwidth]{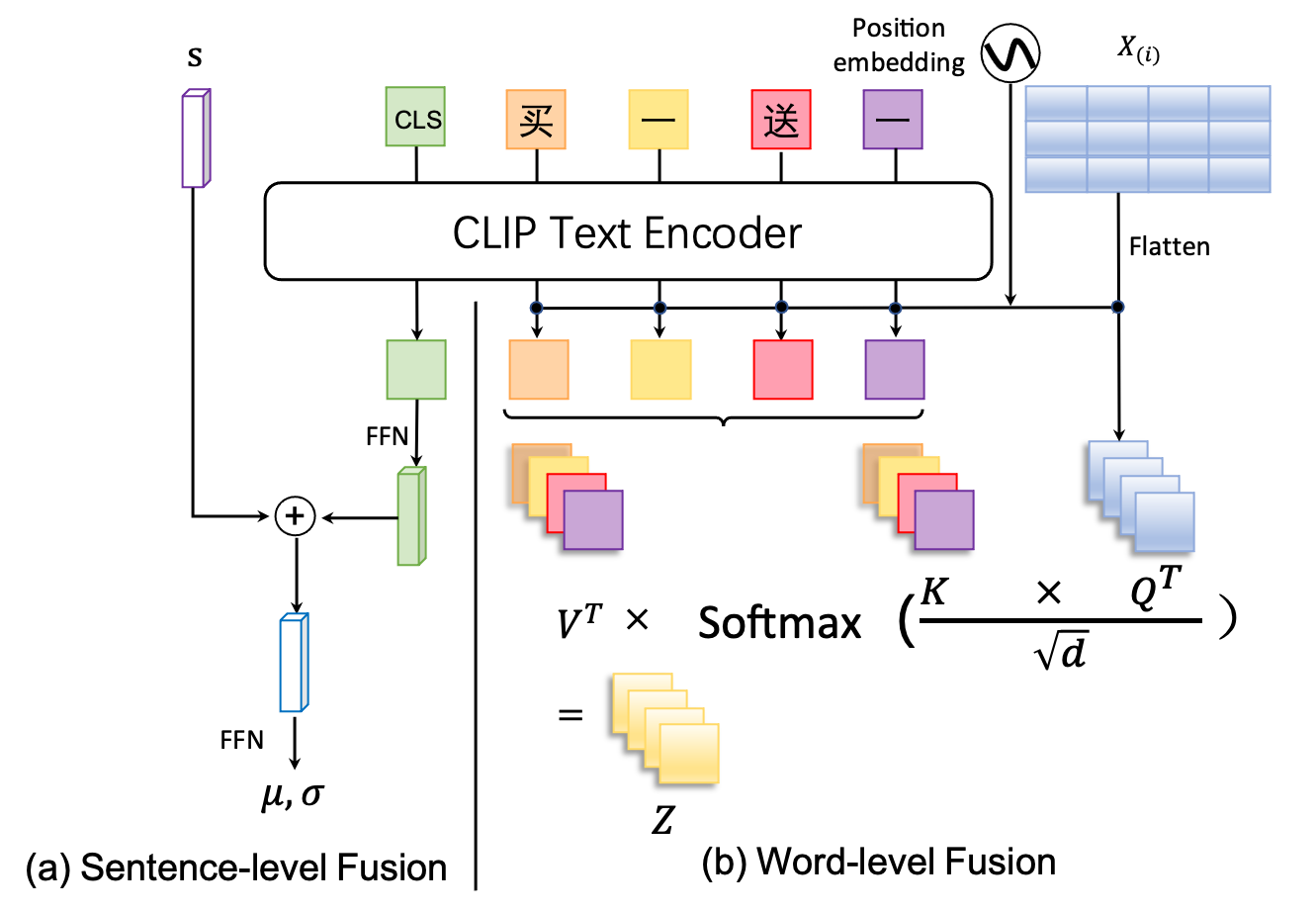}
    \setlength{\abovecaptionskip}{0.cm}
  \caption{Visualization of Text comprehension module. (a) Sentence-level fusion (b) Word-level fusion.}
  \label{fig:attention}
  \vspace{-0.4cm}
\end{figure}
\subsection{Visual Design through Text Comprehension}
Text images consist of two modalities, visual and linguistic, which means that designing visually appealing text on posters requires an understanding of the text's content. To overcome this challenge, TextPainter leverages pre-trained vision-language models such as Chinese-CLIP~\cite{chinese-clip}, which is the Chinese version of CLIP, to extract semantic features from the texts. These features are then merged with visual features. The extracted semantic features include both sentence-level and word/character-level information. To process these features, we propose two novel fusion methods.

\subsubsection{Sentence-level Fusion}
The semantics of a sentence partly reflect the color style of the text image, making it suitable for integration into the generator via AdaIN. The complete semantic feature of the sentence is encoded by the first token $z_0$, after the text content $c$, using Chinese-CLIP. This token $z_0$ is then mapped and added to the style vector $s$ to merge the features of both visual and linguistic modalities.

\subsubsection{Word-level Fusion}
Highlighting the key points is the most crucial principle in poster visual design. Our approach involves utilizing word/character-level semantics to emphasize text keywords, akin to the attention mechanism. Thus, we propose the Semantic-Aware Cross-Attention plugged into the generator to align word/character-level semantic features with visual features. To the best of our knowledge, this is the first instance of utilizing text semantics to aid in generating text images.

Specifically, as shown in Figure~\ref{fig:attention}, the text semantic tokens output from CLIP is denoted by  $Z_t \in \mathbb{R}^{N_t \times C_1}$. Similarly, the visual tokens $Z_v^{(i)}\in \mathbb{R}^{N_i \times C_2}$ reshaped from feature map in the $i$-th generator layer, denoted as $X^{(i)}$.  Next, take visual tokens as query and text semantic tokens as key and value, as shown in Eq \eqref{ep:at1}.
\begin{equation}
Q = Z_v^{(i)} W_v,\quad K = Z_t W_t,\quad V = Z_t W_t \label{ep:at1}
\end{equation}
Where both $W_v \in \mathbb{R}^{C_2 \times d_{k}}$ and $W_t \in \mathbb{R}^{C_1 \times d_{k}}$ represent linear layers that project query, key, and value to the same dimension.

Then, use the attention\cite{AshishVaswani2017AttentionIA} calculation that takes both the text semantic and visual tokens, as shown in Eq \eqref{ep:at2}. 

\begin{equation}
Z = Softmax(\frac{QK^{T}}{\sqrt{d_k}})V + Q \label{ep:at2}
\end{equation}
After that, $Z \in \mathbb{R}^{N_i \times d_k}$ is projected back to the dimension of  the previous visual tokens $Z_v^{(i)}$, as shown in Eq \eqref{ep:at3}
\begin{equation}
Z_{att}^{(i)} = Z W_{out} \label{ep:at3}
\end{equation}
Where $W_{out} \in \mathbb{R}^{d_k \times C_2}$. Finally, the result $Z_{att}^{(i)} \in \mathbb{R}^{N_i \times C_2}$ is reshaped back to $X_{att}^{(i)} \in \mathbb{R}^{H_i \times W_i  \times C_2}$, as the input of next generator layer. 

\label{sec:saa}
Consequently, the attention calculation between character/word-level text semantics and visual features heightens the visibility of keywords in text images.

\subsection{Training Strategies}
\subsubsection{Contextual padding}
Generating text images during the training stage presents a practical challenge as texts have varying lengths, and the sizes of text fonts differ. This renders the implementation of mini-batch training infeasible. Therefore, it is essential to find a solution that can handle text images of varying sizes without distorting the font glyph.

Padding appears to be a better alternative to resizing text images to the same size during training, which is infeasible due to varying text lengths and image sizes. However, filling all images with the same pixel value leads to blurred text image edges, as discovered through extensive experiments. Fortunately, this issue can be resolved by using the background values surrounding the text images for padding, which helps overcome size constraints during training.

\subsubsection{Loss function}
The traditional pixel-wise supervised approach is impractical due to the challenge of annotating text masks. Thus, we adopt a weakly supervised method that utilizes text content and bounding boxes of texts in our training process. Furthermore, we employ adversarial learning to enhance the generator training.

The following loss functions are applied to efficiently train our model to generate text images with harmonious colors.

First, the reconstruction loss $\mathcal{L}_{rec}$ is defined as the L1 norm. This ensures finer details in the generated images.
\begin{equation}
    \mathcal{L}_{rec}= \mathbb{E}\left[\frac{1}{N_t}\left\|I_{t} - O_{t} \right\|_1\right] \label{eq:rec}
\end{equation}
Where $N_t = H_t \times W_t $, which is used as normalization due to the different sizes of text images in our dataset. 

Additionally, in an effort to ensure that the generated image and ground truth possess the same style,the perceptual loss $\mathcal{L}_{per}$ is also utilized during training.
\begin{equation}
    \mathcal{L}_{pre} = \mathbb{E}\left[\sum \limits_{i} \frac{1}{M_i} \left\| \phi_i(I_{t}) -\phi_i(O_{t}) \right\|_1 \right] \label{eq:pre}
\end{equation}
Here, $\phi_i$ is denoted as the feature map of $i$-th layer in ~VGG\cite{KarenSimonyan2014VeryDC} and $M_i = H_i \times H_i \times C_i $.  

Moreover, the adversarial loss is introduced as shown in Eq \eqref{eq:adv}, where $D$ denotes the discriminator that discriminates between the generated image $O_t$ and the real image $I_t$.
\begin{equation}
    \mathcal{L}_{adv} = \mathbb{E}\left[ D(O_t) \right] 
    \label{eq:adv}
\end{equation}

Finally, our model can be jointly optimized based on Eq \eqref{eq:all}.
\begin{equation}
  \mathcal{L} =  \lambda_1 \mathcal{L}_{rec} + \lambda_2 \mathcal{L}_{pre} +\lambda_3 \mathcal{L}_{adv} \label{eq:all}
\end{equation}

\subsubsection{Loss weighting}
GAN training is often plagued by instability, with the discriminator dominating the learning process at the start. To address this issue, greater weight is assigned to the reconstruction loss during early training, and the dynamic adjustment is made based on the following strategy as the generator enhances. 
\begin{equation}
    \lambda_1 = {r}^n 
    \label{eq:loss_weight1}
\end{equation}
\begin{equation}
    \lambda_3 = 1-\lambda_1
    \label{eq:loss_weight2}
\end{equation}
Where $n$ denotes the number of the current training epoch and ${r}$ represents the rate of ${L}_{rec}$, set to 0.85 in our experiments. This method aims to stabilize training and prevent the discriminator from dominating the learning process.

\begin{figure*}[h]
  \centering
  \begin{subfigure}{\linewidth}
    \centering
    \includegraphics[width=\textwidth]{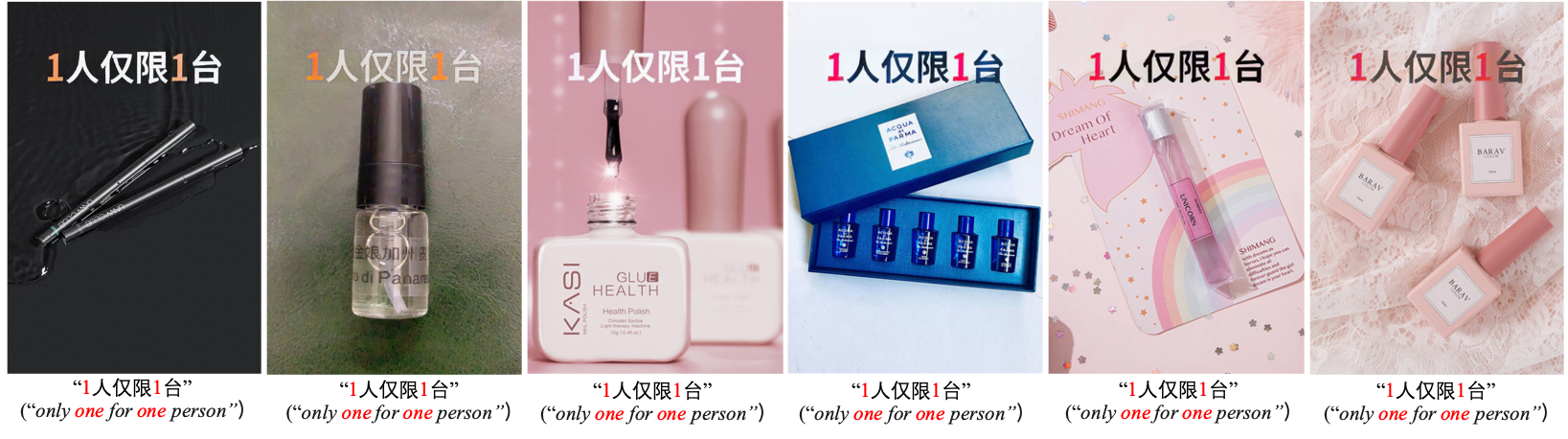}
    \label{fig:ContextAware}
    \vspace{-0.4cm}
  \end{subfigure}
  \hfill
  \begin{subfigure}{\linewidth}
  \centering
  \includegraphics[width=\textwidth]{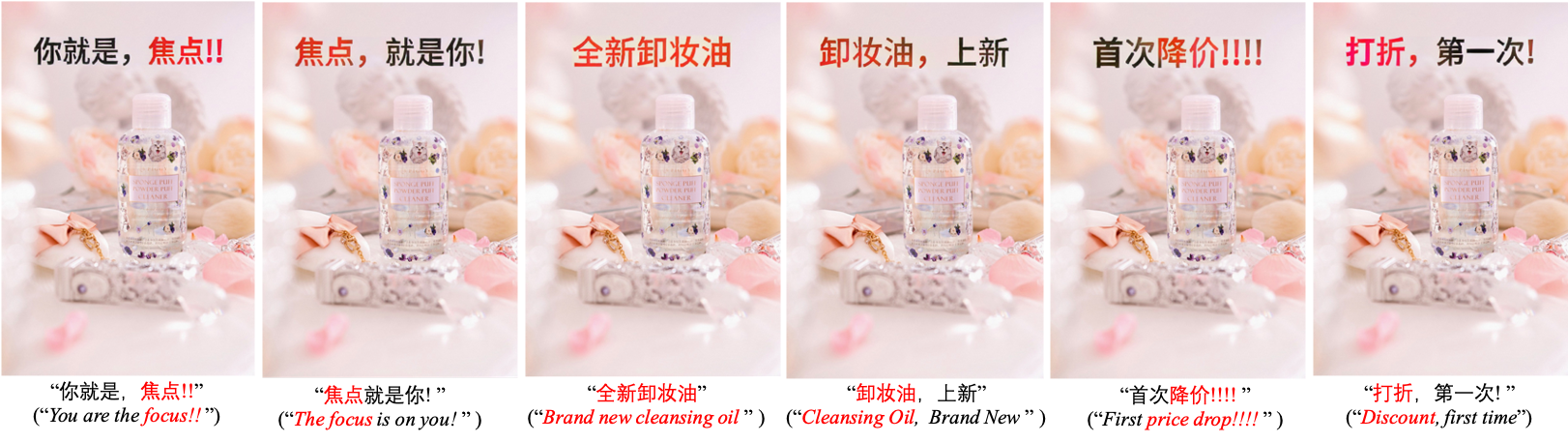} 
    \label{fig:ContentAware}
    \vspace{-0.4cm}
  \end{subfigure}
  
   \setlength{\abovecaptionskip}{0.cm}
  \caption{Generated text images~(paste back to the background images) using TextPainter. 
  (Top) The different results generated by TextPainter given different contextual background images.
  (Bottom) Adaptive change results generated by TextPainter based on different text contents.}
  \label{fig:short}
  \vspace{-0.4cm}
\end{figure*}

\begin{figure}[H]
  \setlength{\abovecaptionskip}{0.cm}
  \includegraphics[width=0.5\textwidth]{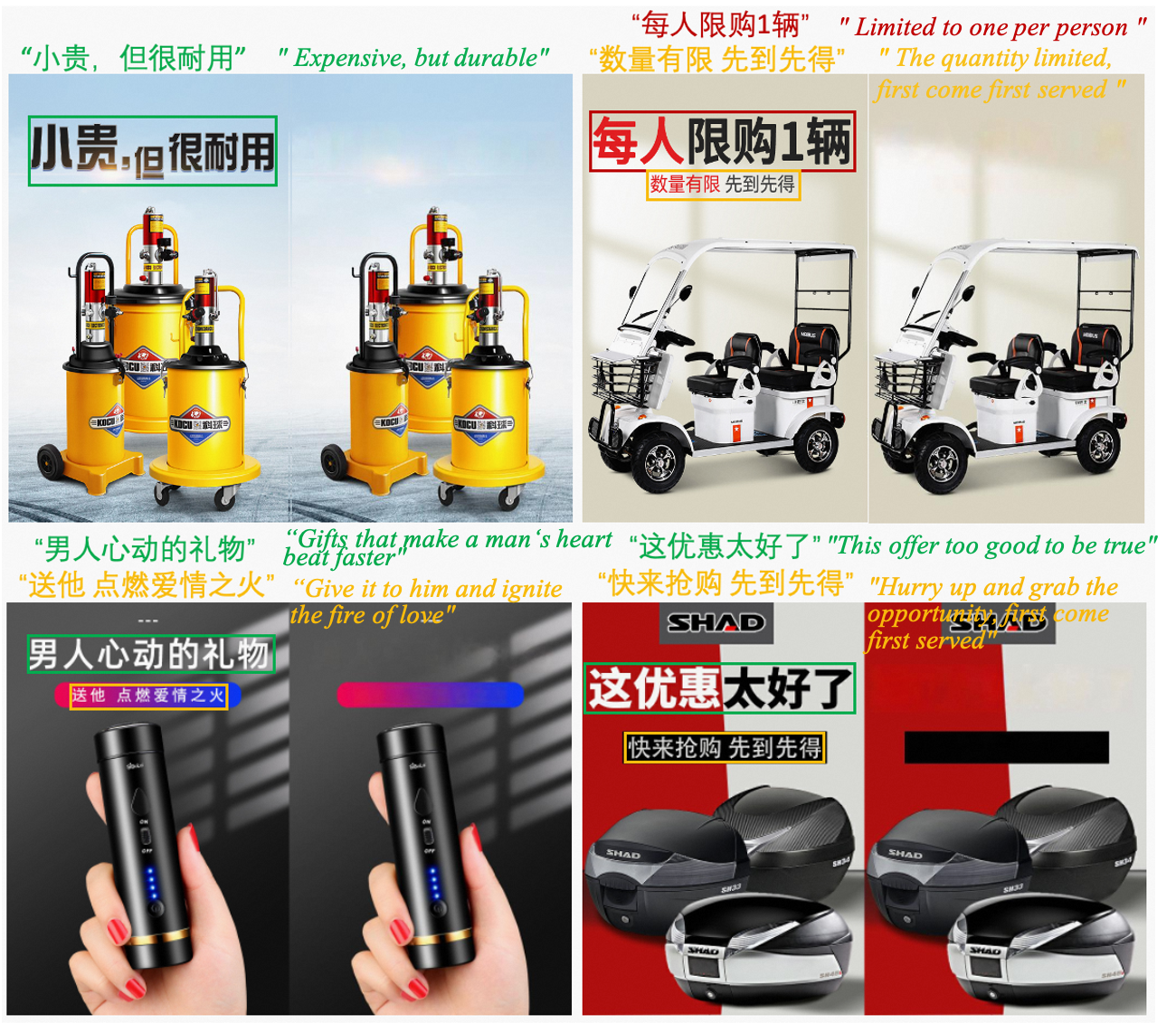}
  \caption{A partial examples of the PosterT80K dataset. }
  \label{fig:dataset}
  \vspace{-0.4cm}
\end{figure}

\section{Poster-T80K Dataset}
Posters play a crucial role in effectively conveying information and often exhibit visually-rich styles. To aid in training and testing the TextPainter network, we collected poster images from Chinese shopping sites, each with a resolution of 513x750. Images that lacked copywriting were filtered out. We collected 165,494 images and filtered out the images without text leaving 117,624 images. The training set and test set are split into 106,009 and 11,615 respectively. The collected posters are annotated with sentence-level bounding boxes and the text string content of each sentence. In order to remove the text from the image and preserve the background, given the labeled text box, we use the text erasing method~\cite{jiang2022self} to perform the erasure of the text on the image.

Figure~\ref{fig:dataset} shows a sample set of poster images in the dataset. Since the poster images and text are manually designed, unlike synthetic text image data usually used for font generation, Poster-T80K better reflects the challenges of real-world text image generation (e.g. irregular gradients and keyword highlight). And Chinese characters have a complex glyph structure compared to English, compared with Book Cover Dataset~\cite{RoxzanneVanEyk2008JudgingAB}.

\vspace{-0.2cm}
\section{Experiments}
\subsection{Implementation Details}
Based on StyleGAN, TextPainter consists of the generator, the glyph encoder, the style encoder, and fusion modules. The glyph and style encoders are implemented using ResNet-34~\cite{KaimingHe2015DeepRL}. The word/character-level fusion module is only plugged into the last two layers of the generator. The text encoder is loaded from the base version of Chinese-CLIP and it performs character-level tokenization for the sentence. After encoding the text, only the first 16 tokens of the token sequence are utilized as input, which is due to the maximum length of the texts in our dataset. 
\vspace{-0.2cm}

\subsection{Evaluation Metrics}
Three commonly-used metrics are adopted for image generation to evaluate the performance of our model: Frechet Inception Distance (FID)\cite{DBLP:conf/nips/HeuselRUNH17}, structural similarity index measure (SSIM)\cite{DBLP:journals/tip/WangBSS04}, Peak signal-to-noise ratio (PSNR). The ground-truth text images and the generated results are compared to calculate these metrics. 

\subsection{Comparsion results}
\paragraph{\textbf{Baseline methods.}}
To the best of our knowledge, no prior work has directly addressed our task. Therefore, a modified version of WebFont~\cite{zhao2018modeling} is our initial baseline approach.  Moreover, we also implement two traditional methods as our baselines. The details of these baselines will be elaborated below. 
\begin{itemize}
    \item \textbf{Base on the classification.} WebFont~\cite{zhao2018modeling} is a classification-based approach that uses information such as images to predict text attributes, which is implemented by ourselves.
    \item \textbf{Base on the color contrast.} The approach extracts the main colors respectively for both local and global images. Then, the color with the highest contrast between the global and local ones is selected from the global main colors.
    \item \textbf{Base on the retrieval.} This method utilizes color histograms extracted from both global and local background images of the poster as features, with text color as the label.
\end{itemize}
\paragraph{\textbf{Baseline Result.}}
Table \ref{tab:baseline} gives the quantitative comparison results of the different methods and on the test set of the dataset that we proposed. The experimental results indicate that the color contrast-based method produces the lowest quantitative results on the test data, possibly because this method relies on human observation-based rules, instead of learning patterns from data. The classification-based methods also show poor performance, as such methods require quantification of the RGB values of colors, which often results in uneven distribution of colors in the color space, making the classification results easily biased towards specific categories. Both retrieval-based methods and TextPainter have obtained favorable quantitative outcomes. Nonetheless, retrieval-based methods necessitate the annotation of text colors for training, whereas TextPainter solely requires bounding boxes. Besides, TextPainter is a GAN-based method with a disadvantage in the FID compared to other methods that are based on graphic rendering.  This issue is proven in our result, therefore TextPainter achieves better results in terms of SSMI and PSNR. Moreover, as depicted in Figure~\ref{fig:short}, our method is based on text comprehension, which is capable of highlighting the keywords in the text. Overall, our method's primary advantage is its capability to infer visually emphasized content from the text semantic.

\begin{table}
  \centering  
    \setlength{\abovecaptionskip}{0.1cm}
  \caption{
  The quantitative results of different baseline methods and TextPainter.
}
  \begin{tabular}{@{}lccc@{}}
    \toprule
    Method & FID$\downarrow$ & SSIM$\uparrow$ & PSNR$\uparrow$ \\
    \midrule
    Classification~\cite{zhao2018modeling} & 23.32 & 0.6801 & 32.46 \\
    Color Contrast & 25.43 & 0.6939 & 32.78\\
    Retrieval & \textbf{17.38} & 0.6928 & 32.78\\
    Ours & \underline{18.53} & \textbf{0.7042} & \textbf{33.49}\\
    \bottomrule
  \end{tabular}
    \vspace{-0.4cm}
  \label{tab:baseline}
\end{table}

\subsection{Analysis}
\label{sec:analysis}
\paragraph{\textbf{Visual context-aware capability.}}
We investigate the capability of TextPainter to comprehend visual contexts by utilizing different poster backgrounds as the style encoder of the input while maintaining the text input as "only one for one person."
In Figure \ref{fig:short} Top, the results illustrate that TextPainter can detect contextual changes and produce text visuals with harmonized colors, leveraging the capabilities of our style encoder.

\paragraph{\textbf{Text semantics-aware capability.}}
In addition, we explore whether TextPainter can comprehend text semantics by fixing the poster background and text position, while varying the text content. As shown in Figure \ref{fig:short}  (Bottom), the outcomes reveal that TextPainter can emphasize distinct keywords in the generated text visuals for different text contents. It is noteworthy that TextPainter is capable of understanding word order~(1st and 2nd columns, 3rd and 4th columns) and synonym substitutions~(the last two columns).

\begin{figure}
   \centering
  \includegraphics[width=0.75\linewidth]{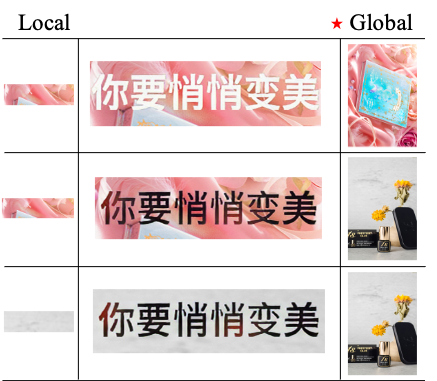}
    \setlength{\abovecaptionskip}{0.1cm}
  \caption{The qualitative result of the color style encoder: "Global" means the poster background input in the color style encoder. "Local" represents the local background used by the input generator for background blending.}
  \label{fig:Effectiveness}
  \vspace{-0.2cm}
\end{figure}

\paragraph{\textbf{Effectiveness of style encoder.}}
The style encoder is utilized to extract the color style of the poster background. However, as shown in the generation block in Figure ~\ref{fig:main_img}, in order to  generate text in the background of the text area, the local background $I_lb$ and feature map are concatenated. This raises concern that the style encoder may fail to function since the generator may directly learn the color style from the local background $I_lb$ and disregard the input from the style encoder. To address this issue, we conducted experiments to validate our concern, as illustrated in Figure~\ref{fig:Effectiveness}. The qualitative results indicated that the color style of the text image is not affected by the local background $I_lb$, but rather by the input from the style encoder. This is due to the fact that TextPaint only blends the text and background in the final block, where the color style of the text image is already established in the previous blocks.

\begin{figure}
   \centering
  \includegraphics[width=0.8\linewidth]{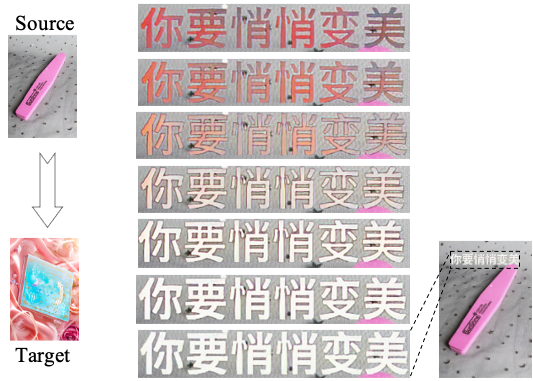}
      \setlength{\abovecaptionskip}{0.1cm}
  \caption{The result of the interpolation experiment where the color style vector is linearly interpolated between the source and target.}
  \label{fig:interpolation}
  \vspace{-0.4cm}
\end{figure}

\paragraph{\textbf{Color style interpolation.}}
In order to investigate additional properties of TextPainter, we performed interpolation experiments on styles extracted by the style encoder. We utilized two different styles extracted from two distinct poster backgrounds for linear interpolation, which served as input to the generator. Figure~\ref{fig:interpolation} demonstrates that the color of the text image can transition smoothly between the two styles. The findings of this study demonstrate that the TextPainter color style exhibits significant continuity in the feature space. Consequently, there is an opportunity to further explore color editing techniques in future research.

\begin{figure*}
  \centering
   \setlength{\abovecaptionskip}{0.cm}
  \includegraphics[width=\textwidth]{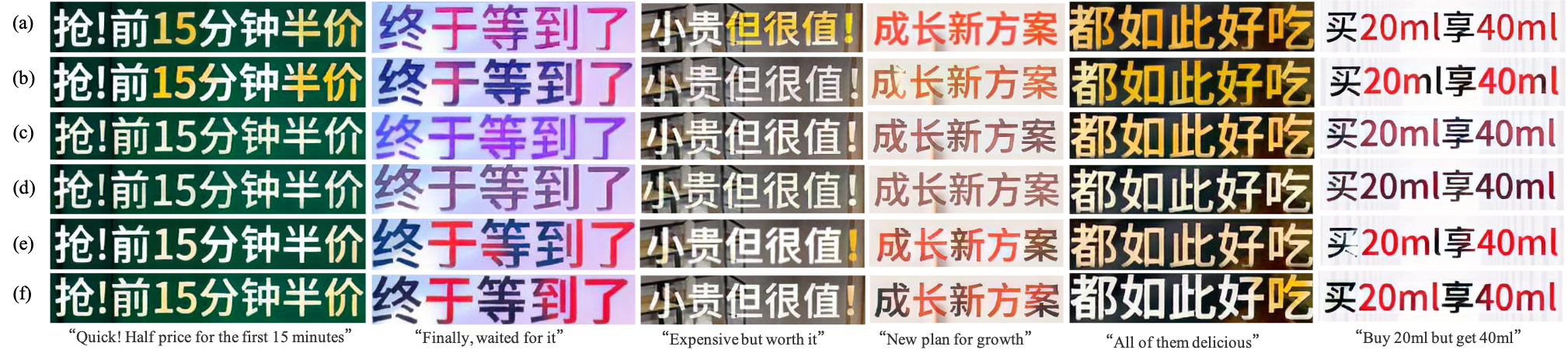}
  \caption{Qualitative ablation experiment results of TextPainter's different modules. (a) Ours. (b) w/o$\,$ Sentence-level Fusion. (c) w/o$\,$ Word-level Fusion. (d) w/o$\,$ Text Encoder. (e) w/o$\,$ Mask.  (f) w/o$\,$ Style Encoder.}
  \label{fig:exp1}
  \vspace{-0.4cm}
\end{figure*}

\paragraph{\textbf{The impact of the artifact after text erasing.}}
The proposed dataset comprises pairs of a poster image ground truth and a background image. 
In particular, the poster background image is generated by the inpainting model~\cite{jiang2022self} to erase the text image in the poster, which may result in artifacts that are not visible to the naked eye and information leakage. To assess the impact of artifacts, a small subset of clean poster backgrounds without artifacts have been collected. For the purpose of comparison, we also created the same background images with artifacts. Finally, compare the results of generating using the two different background images. As shown in Table 1, the qualitative results in Figure~\ref{fig:exp1} indicate that the effect of artifacts is extremely slight.

\begin{table}[H]   
  \centering
  \setlength{\abovecaptionskip}{0.1cm}
  \caption{
  The  quantitative results of test data with artifact and test data without artifact.
}
  \begin{tabular}{@{}lccc@{}}
    \toprule
    Method & FID$\downarrow$ & SSIM$\uparrow$ & PSNR$\uparrow$ \\
    \midrule
    w/o artifact  & 21.49 & 0.6943 & 32.54\\
    w/ artifact & 21.23 & 0.6948 & 32.81 \\
    \bottomrule
  \end{tabular}

  \label{tab:artifact }
    \vspace{-0.5cm}
\end{table}
\subsection{Ablation Studies}
Table \ref{tab:semantic} and Figure \ref{fig:exp1} provide an ablation study evaluating the effects of the color style encoder, text encoder, position mask, and the fusion of text semantics at different granularities, respectively. 
The term "w/o$\,$ Style Encoder" refers to the absence of a color style encoder. Similarly, "w/o$\,$ Mask" denotes the omission of the position mask of text, whereas "w/o$\,$ Text Encoder" signifies the non-existence of a text comprehension module. Additionally, "w/o$\,$ Sentence-level Fusion" denotes the lack of cross-attention, and "w/o$\,$ Word-level Fusion" implies the absence of sentence-level tokens.
\paragraph{\textbf{Color style encoder}}
Based on the result, the ablation of the color style encoder resulted in a noteworthy decline in performance. The main reason is that the absence of a style color encoder model makes it impossible to extract the color style from the visual context of the poster and guide the generation of globally and locally harmonious color text images.
\paragraph{\textbf{Position mask}}
Obviously, as shown in Figure~\ref{fig:exp1}, the lack of masks makes the style encoder only focus on the global background, which can lead to poor color harmony between the text image and the local background.
\paragraph{\textbf{Text Encoder}}
The removal of the text understanding module, either partially or completely, significantly weakens the emphasis on keywords in the text, as illustrated in (b)(c)(d) of Figure~\ref{fig:exp1}. This finding serves as evidence that the inclusion of the text understanding module can markedly enhance our method.


\begin{table}   
  \centering 
  \setlength{\abovecaptionskip}{0.1cm}
  \caption{
  The quantitative results of ablation studies.
}
  \begin{tabular}{@{}lccc@{}}
    \toprule
    Method & FID$\downarrow$ & SSIM$\uparrow$ & PSNR$\uparrow$ \\
    \midrule
    w/o$\,$ Style Encoder & 25.76 & 0.6747 & 32.20 \\
    w/o$\,$ Mask & 19.45 & 0.7011 & 33.01\\
    w/o$\,$ Text Encoder & 20.52 & 0.6839 & 32.42\\
    w/o$\,$ Sentence-level Fusion & 18.98 & 0.7056 & 33.28\\
    w/o$\,$ Word-level Fusion & 20.21 & 0.6991 & 32.87\\
    Ours & \textbf{18.53} & \textbf{0.7042} & \textbf{33.49} \\
    \bottomrule
  \end{tabular}

  \label{tab:semantic}
  \vspace{-0.4cm}
\end{table}


\subsection{User study}
\begin{figure}[H]
 \centering
  \includegraphics[width=0.8\linewidth]{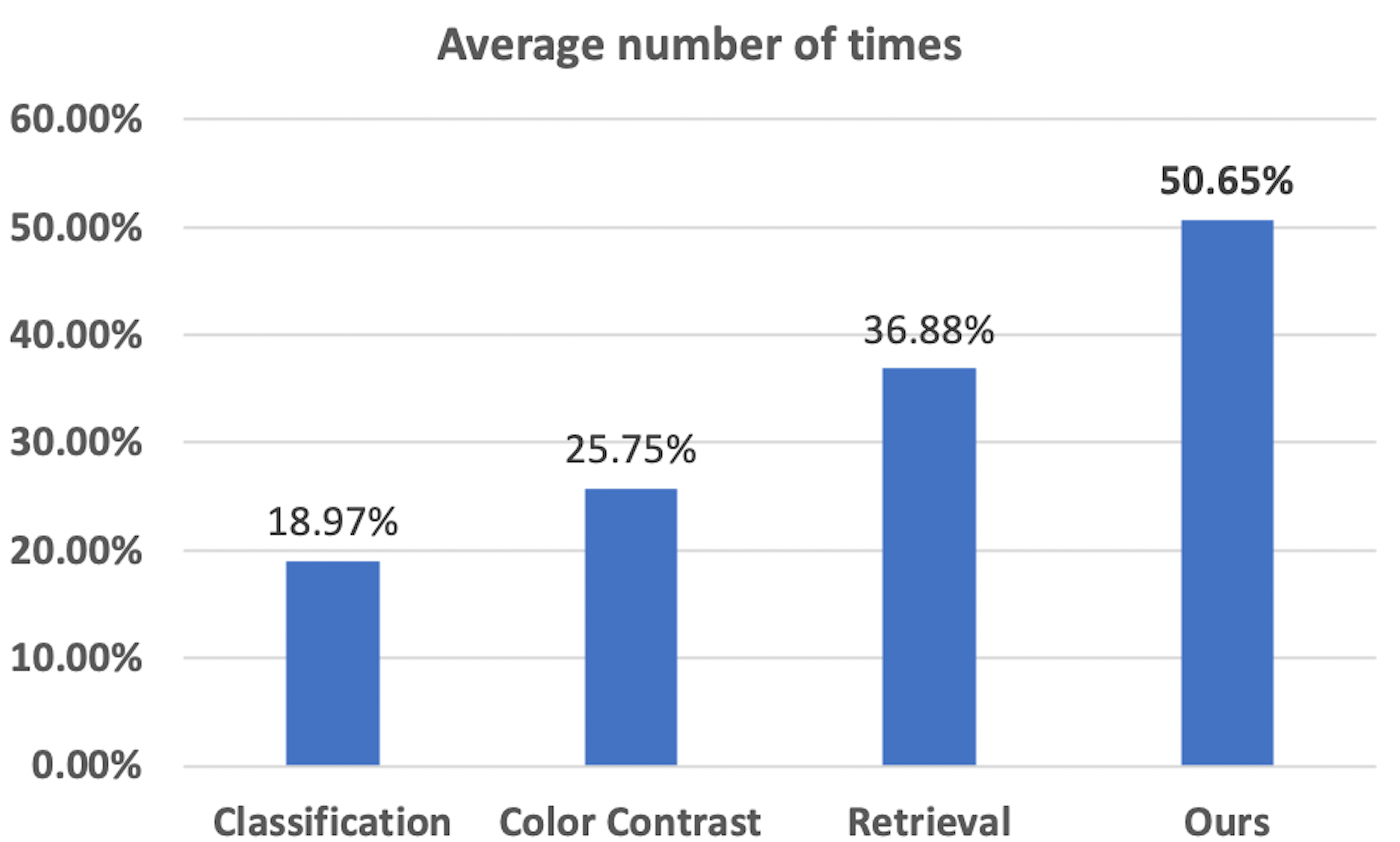}
      \setlength{\abovecaptionskip}{0.2cm}
  \caption{The User study results on the aesthetics of different methods.}
  \label{fig:user_study}
  \vspace{-0.4cm}
\end{figure}

The evaluation  of poster design emphasizes aesthetics. Due to the difficulty of quantifying aesthetics, we adopted a user study to evaluate different methods. 
Specifically, we randomly sampled 32 groups of poster images generated from the test results of different method. Then we presented participants with randomly-ordered generated images from distinct methods. 30 users were asked to choose the poster images of the most aesthetic appeal in the group. The aesthetic result of the user study is shown in Figure~\ref{fig:user_study}, where numbers represent the percent of times users chose. According to the results, the highest number of users voted for TextPainter, which indicates that our TextPainter, based on the text understanding module, can automatically generate more attractive posters.



\section{Conclusion}
In this paper, we present the task of text image generation on the poster for the first time. Our method aims to automatically generate clear, harmonious, colorful, and creative text images on posters based on image contextual and text content, like a designer. Experiments validate the ability of our method to model contextual visual and textual semantics and to generate visually appealing and meaningful text images. The extent experiment demonstrates the potential of the method in scalable training through a self-supervised training approach. Finally, we collect a novel Chinese poster dataset with annotation and hope our work will encourage future research on multi-modal text image generation.

\begin{acks}
This work is supported by Alibaba Group through Alibaba Innovation Research Program, the National Nature Science Foundation of China (62121002, 62272436), the Fundamental Research Funds for the Central Universities under Grant WK3480000011, the MCT Key Lab of CCCD, the Fundamental Research Funds for the Central Universities WK2100000026, the Anhui Provincial Natural Science Foundation 2208085QF190.
\end{acks}

\clearpage
\bibliographystyle{ACM-Reference-Format}
\balance
\bibliography{sample-sigconf}

\clearpage
\appendix

\section{PosterT80K Dataset}
This section provides a comprehensive introduction to the PosterT80K dataset. To begin with, we present a statistical analysis of the dataset. Next, we describe the data processing methods that we employed. The raw dataset comprises 117,624 Chinese poster images that we collected from e-commerce websites. Each poster image has a size of 513x750 and contains multiple texts, which total 376,844 text images. We carefully labeled the content and position of each text image by manual annotation. Specifically, we denoted the content as $c$ and the position as $p=(x, y, w, h)$. After dataset processing, the training set contains 106,009 posters and 148,891 text images, while the testing set comprises 11,615 posters and 16,603 text images.

\subsection{Data distribution analysis}
\begin{figure}[H]
 \centering
 \setlength{\abovecaptionskip}{0.1cm}
  \includegraphics[width=\linewidth]{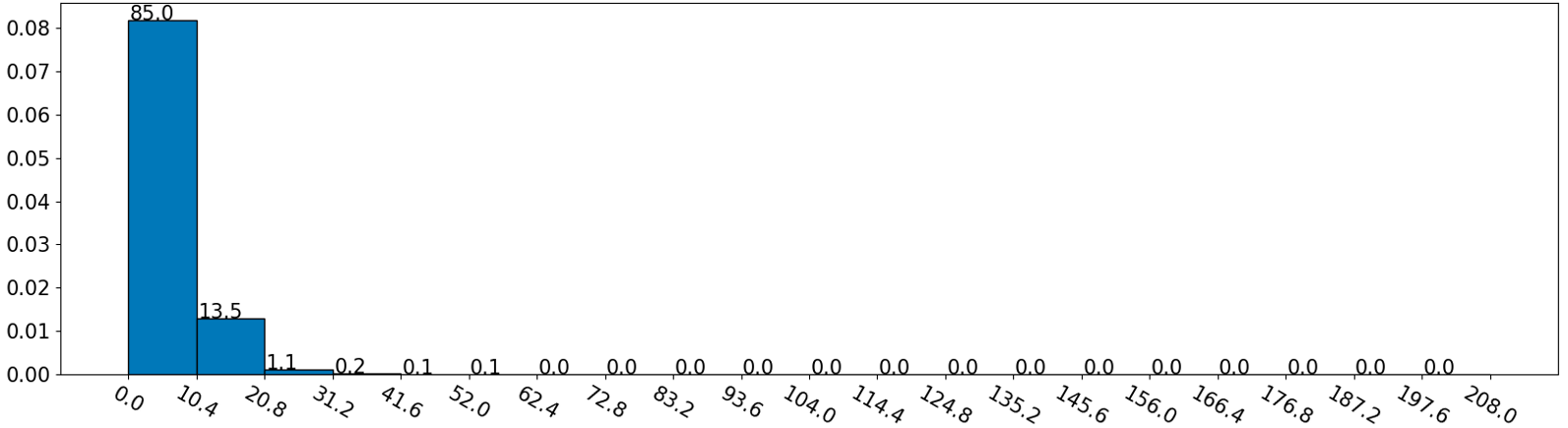}
  \caption{The distribution of text length in the raw dataset. (horizontal axis) text length. (vertical axis) probability density.
}
  \label{fig:len_text}
  \vspace{-0.4cm}
\end{figure}

\begin{figure}[H]
 \centering
  \setlength{\abovecaptionskip}{0.1cm}
  \includegraphics[width=\linewidth]{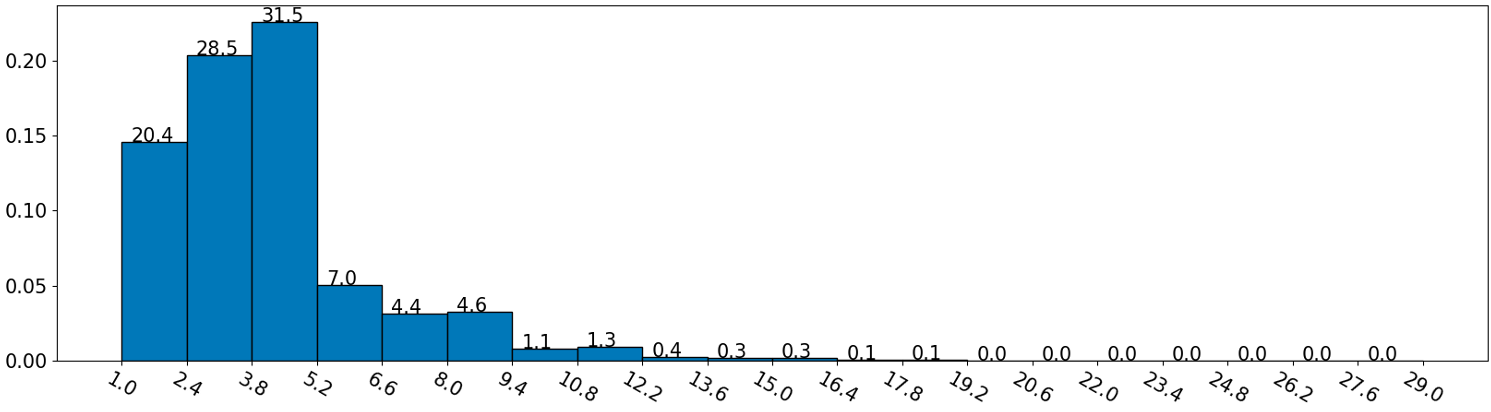}
  \caption{The distribution of the number of text lines that posters contain in the raw dataset.(horizontal axis) the number of text lines. (the vertical axis) the probability density.
}
  \label{fig:num_text}
  \vspace{-0.4cm}
\end{figure}

\begin{CJK*}{UTF8}{gbsn}
\paragraph{\textbf{Text content analysis}}
The distribution of text-related features was the primary focus of our analysis. Figure \ref{fig:len_text} illustrates that nearly 90\% of the text lines in the dataset contain less than 12 characters, while almost 99\% contain less than 22 characters. These findings suggest that most of the text lines in the posters are short sentences. Besides, the top 20 frequently appearing characters are identified and recorded in Table \ref{tab:char_times}. Since e-commerce posters have a specific purpose of highlighting selling points and prices, numbers and some Chinese characters, such as ‘抢’ (rush), ‘立’ (stand), and ‘购’ (buy), frequently feature in the text. Figure \ref{fig:num_text} reveals that approximately 80\% of posters contain 1 to 5 texts, and almost 99\% of posters contain 1 to 12 texts.
\end{CJK*}

\begin{figure}[H]
 \centering
  \setlength{\abovecaptionskip}{0.1cm}
  \includegraphics[width=\linewidth]{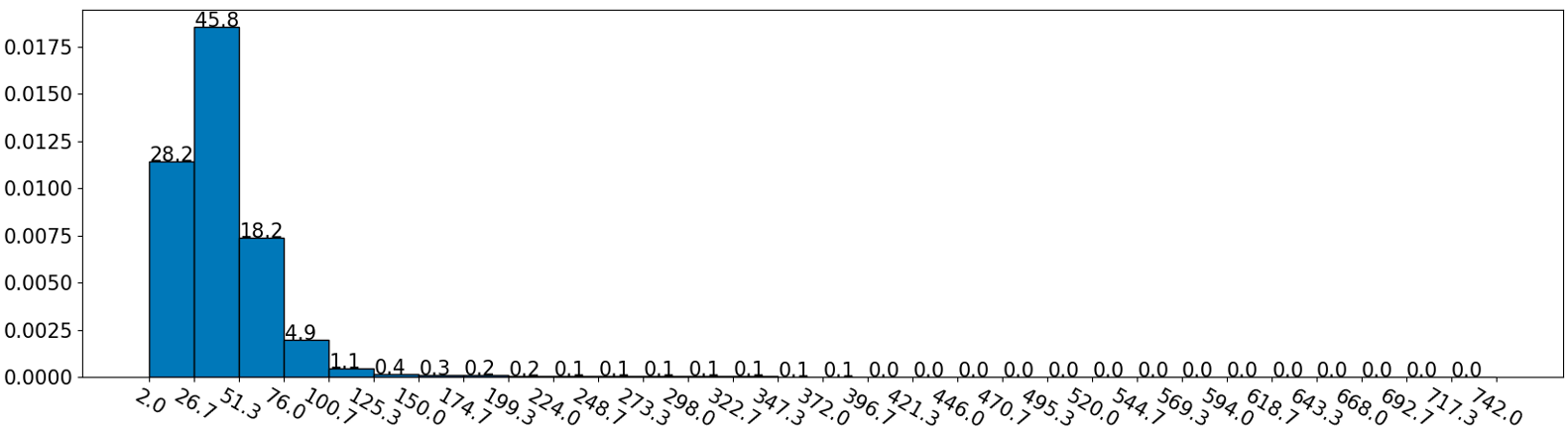}
  \caption{
   The distribution of the height of the text images in the raw dataset. (horizontal axis) the height. (vertical axis) the probability density.
}
  \label{fig:h_dis}
  \vspace{-0.4cm}
\end{figure}

\begin{CJK*}{UTF8}{gbsn}
\begin{table}[H]
 \setlength{\abovecaptionskip}{0.2cm}
\begin{tabular}{|c|c|c|c|c|c|}
\hline
\textbf{}   & \textbf{Character} & \textbf{Count} & \textbf{}   & \textbf{Character} & \textbf{Count} \\ \hline
\textbf{1}  & ‘0’                & 89048          & \textbf{11} & ‘3’                & 25268          \\ \hline
\textbf{2}  & ‘1’                & 78131          & \textbf{12} & ‘即’ (immediately)  & 23265          \\ \hline
\textbf{3}  & ‘ ’                & 65471          & \textbf{13} & ‘元’ (yuan)         & 20906          \\ \hline
\textbf{4}  & ‘2’                & 41279          & \textbf{14} & ‘一’                & 20123          \\ \hline
\textbf{5}  & ‘9’                & 29215          & \textbf{15} & ‘买’ (buy)          & 19792          \\ \hline
\textbf{6}  & ‘抢’ (rush)         & 28132          & \textbf{16} & ‘价’ (price)        & 18270          \\ \hline
\textbf{7}  & ‘立’ (stand)        & 28035          & \textbf{17} & ‘4’                & 14641          \\ \hline
\textbf{8}  & ‘\textgreater{}’   & 27069          & \textbf{18} & ‘限’ (limit)        & 14599          \\ \hline
\textbf{9}  & ‘购’ (buy)          & 25625          & \textbf{19} & ‘送’ (give)         & 14555          \\ \hline
\textbf{10} & ‘5’                & 25622          & \textbf{20} & ‘新’ (new)          & 14265          \\ \hline
\end{tabular}
\caption{The top 20 characters that appear in the text of the dataset.}
\label{tab:char_times}
\vspace{-0.4cm}
\end{table}
\end{CJK*}

\begin{figure}[H]
 \centering
  \setlength{\abovecaptionskip}{0.cm}
  \includegraphics[width=\linewidth]{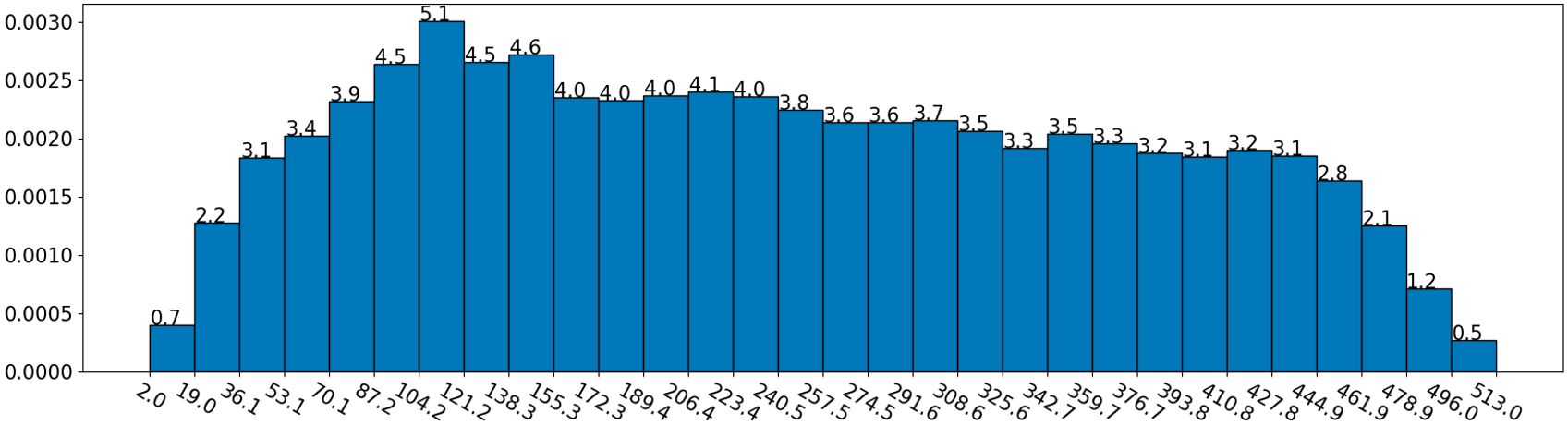}
  \caption{
   The distribution of the width of the text images in the raw dataset, with the horizontal axis representing the width and the vertical axis representing the probability density.
}
  \label{fig:w_dis}
  \vspace{-0.4cm}
\end{figure}

\begin{figure}[H]
 \centering
  \setlength{\abovecaptionskip}{0.1cm}
  \includegraphics[width=\linewidth]{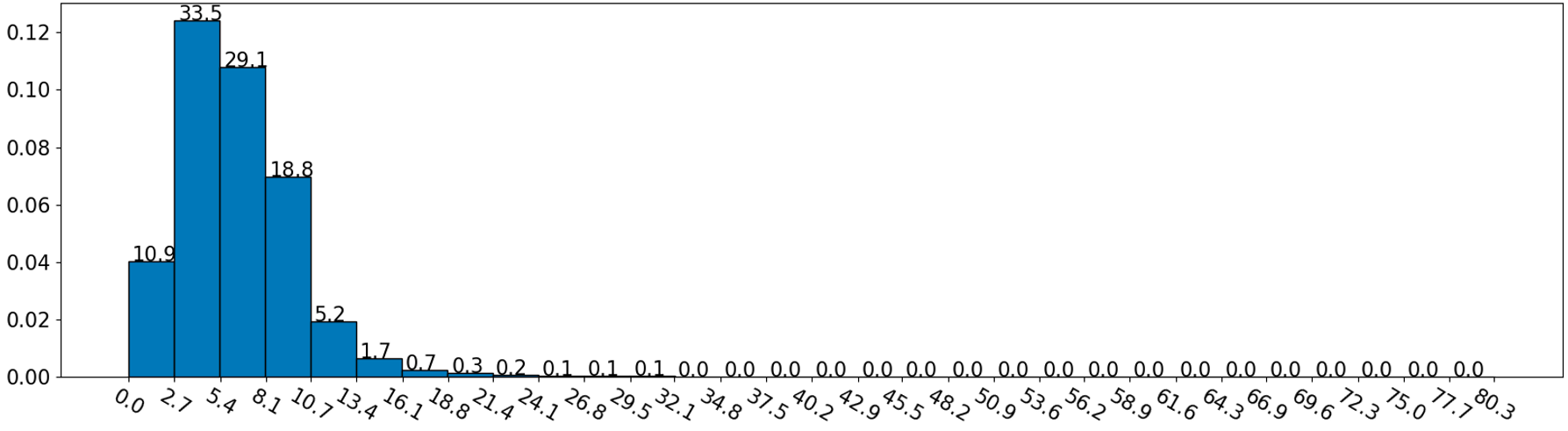}
  \caption{
   The distribution of the aspect ratio of the text images in the raw dataset. (horizontal axis) the aspect ratio. (vertical axis) the probability density.
}
  \label{fig:r_dis}
  \vspace{-0.4cm}
\end{figure}

\paragraph{\textbf{Distribution of text image size}}
Although the size of the image of the poster is fixed, the size of the text image varies, which complicates the training of the model. To address this issue, we performed an analysis of the text image sizes. The distributions of  height and width of the text image are presented in Figure \ref{fig:h_dis} and Figure \ref{fig:w_dis}, respectively. The results show that approximately 94\% of the height values are within the range of 16 to 99, exhibiting a high level of concentration, while the width values display a broader distribution ranging from 25 to 513. Furthermore, as depicted in Figure \ref{fig:r_dis}, the aspect ratio is also highly concentrated, with approximately 95\% of the values falling between 1.0 and 11.0, which is consistent with the distribution of our text length.

\begin{figure}[H]
 \centering
  \setlength{\abovecaptionskip}{0.1cm}
  \includegraphics[width=\linewidth]{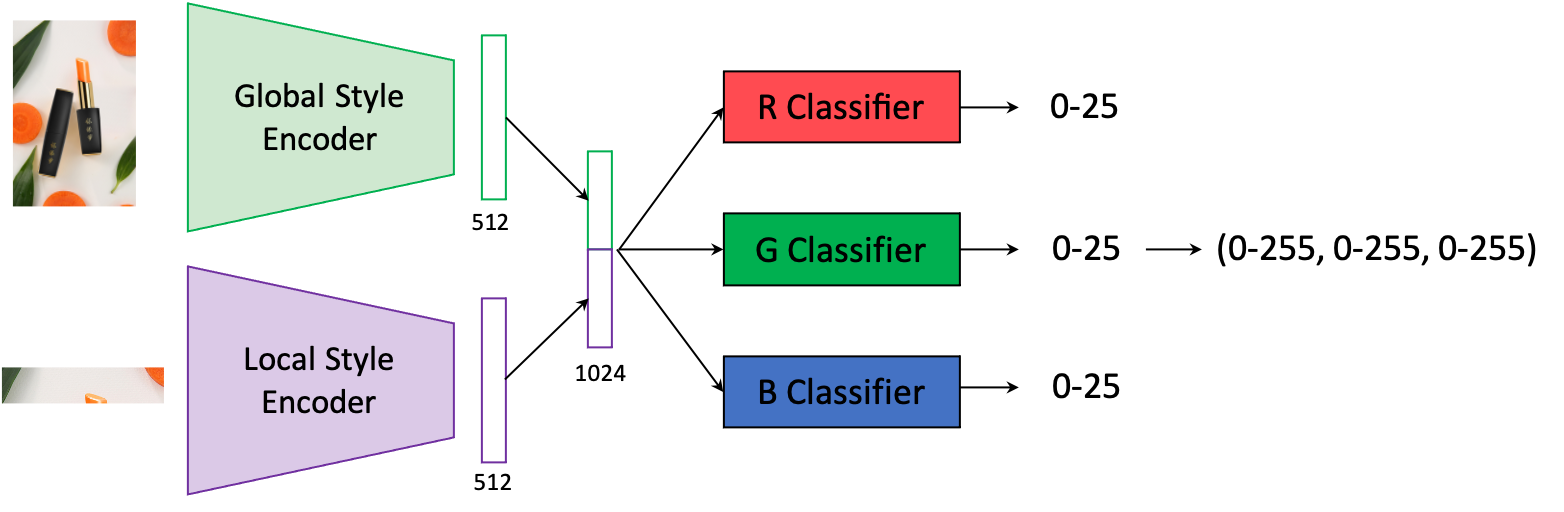}
  \caption{The architecture of the baseline method based on classification.}
  \label{fig:baseline_cls}
  \vspace{-0.4cm}
\end{figure}

\subsection{Data preprocessing}
According to the data distribution analysis, the dataset was first filtered to exclude outlier values in order to facilitate model training, after which the dataset was divided. The filtering process involved the following steps: 
\begin{itemize}
\item Removing posters containing more than 5 text images.
\item Removing text images with aspect ratios greater than 11.
\item Removing text images with heights outside of the range of 30 to 100 or widths outside of the range of 50 to 450.
\item Removing text images outside of the content length range of 1 to 11 characters.
\end{itemize}
Finally, the dataset was then divided based on poster images. 

\section{Model architecture}
The style encoder and glyph encoder of TextPainter are both implemented by ResNet34, with the dimension of the style feature vector being 512.  Besides, the model architectures of the generator are shown in Table \ref{tab:generator}. Furthermore, AdaIN is implemented in StyleConv by a Linear layer and a Conv2DMod layer, while RGBConv is implemented by a StyleConv, an UpSample layer, and a Blur layer, consistent with the implementation of StyleGAN. Finally, we use the base version of Chinese-CLIP text encoder.

\begin{table}
 \setlength{\abovecaptionskip}{0.1cm}
\begin{tabular}{|c|c|c|}
\hline
\textbf{Layer}          & \textbf{Configurations} & \textbf{Output}                                            \\ \hline
Input                   & Glyph Feature Map       & $512\times\frac{H}{32}\times\frac{W}{32}$                  \\ \hline
Conv                    & Conv2D                  & $512\times\frac{H}{32}\times\frac{W}{32}$                  \\ \hline
\multirow{4}{*}{Block1} & StyleConv               & \multirow{4}{*}{$512\times\frac{H}{32}\times\frac{W}{32}$} \\ \cline{2-2}
                        & StyleConv               &                                                            \\ \cline{2-2}
                        & LeakyReLU               &                                                            \\ \cline{2-2}
                        & RGBConv                 &                                                            \\ \hline
\multirow{6}{*}{Block2} & Cross-Attention         & \multirow{6}{*}{$256\times\frac{H}{16}\times\frac{W}{16}$} \\ \cline{2-2}
                        & Upsample                &                                                            \\ \cline{2-2}
                        & StyleConv               &                                                            \\ \cline{2-2}
                        & StyleConv               &                                                            \\ \cline{2-2}
                        & LeakyReLU               &                                                            \\ \cline{2-2}
                        & RGBConv                 &                                                            \\ \hline
\multirow{6}{*}{Block3} & Cross-Attention         & \multirow{6}{*}{$128\times\frac{H}{8}\times\frac{W}{8}$}   \\ \cline{2-2}
                        & Upsample                &                                                            \\ \cline{2-2}
                        & StyleConv               &                                                            \\ \cline{2-2}
                        & StyleConv               &                                                            \\ \cline{2-2}
                        & LeakyReLU               &                                                            \\ \cline{2-2}
                        & RGBConv                 &                                                            \\ \hline
\multirow{6}{*}{Block4} & Cross-Attention         & \multirow{6}{*}{$64\times\frac{H}{4}\times\frac{W}{4}$}    \\ \cline{2-2}
                        & Upsample                &                                                            \\ \cline{2-2}
                        & StyleConv               &                                                            \\ \cline{2-2}
                        & StyleConv               &                                                            \\ \cline{2-2}
                        & LeakyReLU               &                                                            \\ \cline{2-2}
                        & RGBConv                 &                                                            \\ \hline
\multirow{6}{*}{Block5} & Cross-Attention         & \multirow{6}{*}{$32\times\frac{H}{2}\times\frac{W}{2}$}    \\ \cline{2-2}
                        & Upsample                &                                                            \\ \cline{2-2}
                        & StyleConv               &                                                            \\ \cline{2-2}
                        & StyleConv               &                                                            \\ \cline{2-2}
                        & LeakyReLU               &                                                            \\ \cline{2-2}
                        & RGBConv                 &                                                            \\ \hline
\multirow{5}{*}{Block6} & Upsample                & \multirow{5}{*}{$3\times H\times W$}           \\ \cline{2-2}
                        & StyleConv               &                                                            \\ \cline{2-2}
                        & StyleConv               &                                                            \\ \cline{2-2}
                        & LeakyReLU               &                                                            \\ \cline{2-2}
                        & RGBConv                 &                                                            \\ \hline
\end{tabular}
\caption{The architecture of the generator.}
\label{tab:generator}
\vspace{-0.5cm}
\end{table}                                                                                                  

\section{The implementation of baselines}

\subsection{Base on the classification.}
WebFont~\cite{zhao2018modeling} is a classification-based approach that uses information such as images to predict text attributes. As our task only relies on the poster background and text, we implemented WebFont as shown in Figure \ref{fig:baseline_cls}. Specifically, we quantized the RGB color values into 26 categories based on the WebFont settings and then used the global and local style encoders to extract features, which were fused to predict the text color.

\subsection{Base on the color contrast.}
Firstly, extract five theme colors from the overall background of the poster. Secondly, extract one theme color from the local background of the text position. Finally, use the color with the highest contrast with the local theme color among the selected five global theme colors as the text color. The Modified Median Cut Quantization algorithm \cite{bloomberg2008color} is used to extract the theme colors.

\subsection{Base on the retrieval.}
Firstly, RGB color histograms are separately extracted from the global and local backgrounds and concatenated as the text image feature. The histogram dimension for each single color component is 128, and the total feature dimension is 768. Then, the RGB color of the text serves as the label and is used to establish a retrieval database together with the text image feature. During the  inference stage, the same feature extraction method is used, and the most similar sample is retrieved from the database to predict text color.

\begin{figure}[H]
 \centering
  \setlength{\abovecaptionskip}{0.1cm}
  \includegraphics[width=0.65\linewidth]{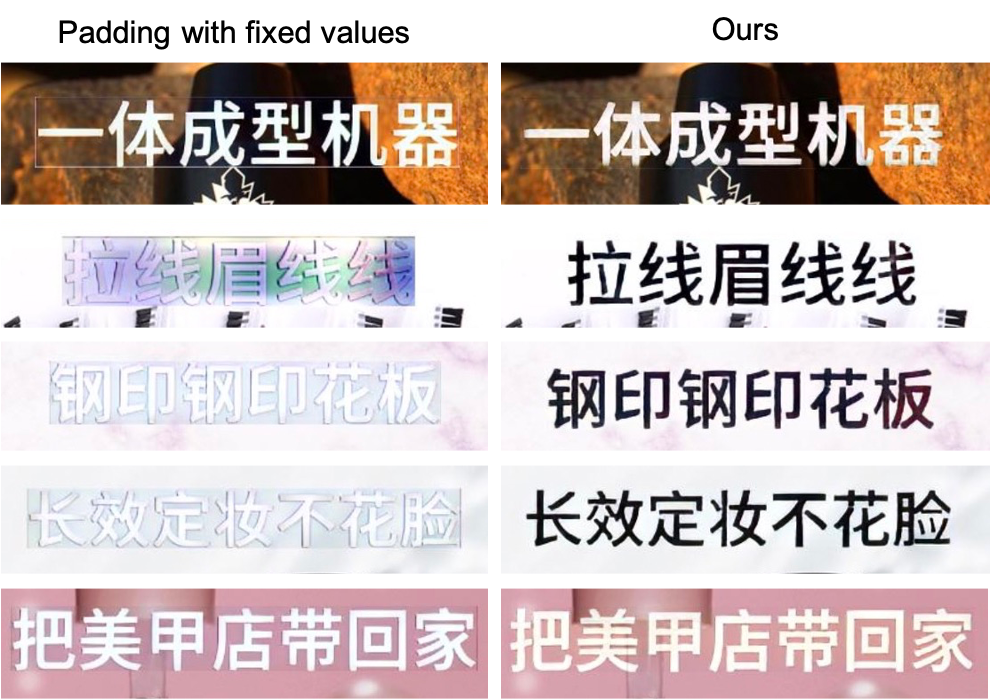}
  \caption{The qualitative results of the impact of the two different padding methods.}
  \label{fig:cmp_padding}
  \vspace{-0.4cm}
\end{figure}

\section{The Contextual Paddding}
The use of padding is necessary to conform text images of varying sizes to a batch and to avoid any negative impact on font size resulting from resizing. However, as shown in Figure \ref{fig:cmp_padding}, our experimentation has revealed that fixed padding values cause distortion in the generated local background. To overcome this issue, we propose an alternative solution that utilizes the background values surrounding the text area for padding. As demonstrated in Figure \ref{fig:cmp_padding}, this approach results in a significant improvement in the overall effect, enabling a better blend between the local and surrounding backgrounds. 

\end{sloppypar}
\end{document}